\pdfoutput=1

\documentclass[11pt]{article}

\usepackage{emnlp2021}

\usepackage{times}
\usepackage{latexsym}

\usepackage[T1]{fontenc}

\usepackage[utf8]{inputenc}

\usepackage{microtype}

\usepackage{dirtytalk}
\usepackage{svg}
\usepackage{xcolor}
\usepackage{hyperref}
\usepackage{amsmath,amssymb,mathtools}
\usepackage{booktabs}

\usepackage[colorinlistoftodos,prependcaption,textsize=scriptsize]{todonotes}

\DeclareMathOperator*{\argmax}{argmax}
\DeclareMathOperator*{\abs}{abs}

\usepackage{graphicx}
\graphicspath{ {./graphics/} }

\title{Memory and Knowledge Augmented Language Models for Inferring Salience in Long-Form Stories}

\author{David Wilmot \\
  School of Informatics\\
  University of Edinburgh \\
  \texttt{david.wilmot@ed.ac.uk}\\
  \And
  Frank Keller \\
  School of Informatics\\
  University of Edinburgh \\
  \texttt{keller@inf.ed.ac.uk}\\
  }

\begin{document}
\maketitle
\begin{abstract}

Measuring event \textit{salience} is essential in the understanding of stories. This paper takes a recent unsupervised method for \textit{salience} detection derived from Barthes Cardinal Functions and theories of surprise and applies it to longer narrative forms. We improve the standard transformer language model by incorporating an external knowledgebase (derived from Retrieval Augmented Generation) and adding a memory mechanism to enhance performance on longer works. We use a novel approach to derive salience annotation using chapter-aligned summaries from the Shmoop corpus for classic literary works.  Our evaluation against this data demonstrates that our \textit{salience} detection model improves performance over and above a non-knowledgebase and memory augmented language model, both of which are crucial to this improvement. 

\end{abstract}

\section{Introduction}

\citet{forster1985aspects} compared a story to a \textit{wriggling worm of time} that can be seen as a series of events arranged in order (see also \citealt{abbott2008cambridge}) --- dinner comes after breakfast,  night after day, nemesis follows hubris.  Not all events are of equal importance, and some far more \textit{salient} than others. For example, the beginning of Dickens' \textit{Great Expectations} ---
\textit{Keep still, you little devil, or I'll cut your throat!} --- is more \textit{salient} to the story than events such as \textit{my sister had a trenchant way of cutting our bread and butter for us}. \textit{Salient} events in storytelling are those that drive the plot forward, change the state in the story world, as opposed to descriptive details or non-consequential activities. As such, detecting \textit{salience} is an essential part of understanding narrative. Detecting salient events has important downstream applications such as summarisation; salient events are the core of plots and can aid storyline writing and story generation; they represent essential information and are relevant to question answering.

This paper builds on the work of \citet{otake-etal-2020-modeling}, who use Barthes Cardinal Functions (BCF) for unsupervised \textit{salience} detection. We augment this approach with a knowledgebase (KB) and memory. Barthes Cardinal Functions \citep{Barthes1966AnIT} are hinge events that cannot be deleted without altering the story; they are the decision points between alternative consequential paths. Barthes and Duisit also define \textit{catalysers}, which are inconsequential events such as the bread and butter example, \textit{indices}, which are descriptive, referring to a character or situation, and \textit{informants}, which identify time and space. These latter types can be seen as \textit{satellites} around the \textit{nuclei}, or filling in gaps between cardinal functions. Hence to identify BCF is to identify the main skeleton of the plot. We treat the BCF events as the salient events in a story. This scheme relates in narratology with \citet{chatman1980story} \textit{kernels} and \textit{satellites} model, as well as with discourse theory in RST \citep{Mann1988RhetoricalST}, which similarly has \textit{nuclei} and \textit{satellites} and more loosely with SDRT \citep{Asher2005LogicsOC} with \textit{coordinating} and \textit{subordinating} relations. The key to the Otake et al. method is that it can be implemented using any LM (Language Model) on any text and does not require a large corpus of annotated training data. 


In this paper, we extend the BCF concept by exploring new measures of salience derived from structural manipulations: We infer \textit{swap salience}, which is swapping rather than deleting an event within the BCF framework. \citet{schmid2003narrativity} discusses how an event can be salient if a reader expects it, but it is unexpected to the character in the story. The reader puts themselves into the character's shoes. \citet{zillmann1996psychology} emphasises how suspense is driven by anticipation and apprehension on behalf of characters the reader cares about. \citet{Bae2009SuspenseSO} propose to use this knowledge disparity between the reader and the character to create more suspenseful plots and hence more important events. We model \textit{knowledge salience} as the difference between an expert-informed reader versus a naive one by taking the difference between the average log-likelihood of a base LM and an LM enriched with memory and a KB. We also take inspiration from the model of \citet{wilmot-keller-2020-modelling}, who compute suspense and surprise in short stories using vector states from a hierarchical model; this follows from theoretical work by \citet{Ely2015SuspenseAS}, and cognitive work from \citet{Li2019TheCM}. We show how a \textit{vector salience} measure can be computed based on this approach.

In addition to exploring new salience measures, our work aims to overcome limitations of existing work on salience modeling. \citet{otake-etal-2020-modeling} only evaluate their model on a single type of narrative (Russian fairytales) and on a very small annotated dataset. We address this by using aligned summaries from the Shmoop corpus \citep{DBLP:journals/corr/abs-1912-13082} to provide salience labels. This results in a large dataset of long works (novels and plays) annotated with silver-standard salience labels. A second limitation of Otake et al. is that they use GPT-2 \citep{radford2019language} as their LM, which has a relatively short context of a few hundred wordpieces. While this works for short stories, the context is too short to track lengthy novels or plays. Often a character will disappear for a long period; for example, \textit{Abel Magwitch} in \textit{Great Expectations}. Plots are often non-linear with recalls and flash-forwards, and the same characters and places reoccur at intermittent points in the story. At any moment in the story, the most relevant passages are not the most recent but the previous actions of the characters, places, and situations involved.

We address this limitation by incorporating an episodic knowledge retrieval mechanism (derived from RAG; \citealt{NEURIPS2020_6b493230}) and fuse this with a short-term memory mechanism that can extend the capabilities of a transformer LM. The intent is that the memory will learn to recall the most relevant parts of the story, act as an implicit index into these dimensions, and the KB will supplement this with typical plot knowledge. This memory mechanism is much more suitable than recent work on extended transformers for longer sequences, see \citet{DBLP:journals/corr/abs-2009-06732} and \citet{DBLP:journals/corr/abs-2103-14636} for thorough reviews. Characters, places, subplots ebb and flow in long stories, so the most relevant information may be hundreds of pages previous with mainly irrelevant information in-between, which suits indexed episodic memory rather than a transformer that must filter out the mainly irrelevant details in-between. For example, \textit{Abel Magwitch} in \textit{Great Expectations} is in the first two chapters and then reappears explicitly in Chapter 40.


Our results show that integrating KB and memory components improves the overall performance of \textit{salience} detection. Using a vector alternative to infer \textit{salience} is a slight improvement over the LM. Other measures such as \textit{swap salience} and \textit{knowledge salience} perform worse than the main salience measures but still show improvements over our baseline model.

\section{Related Work}

The main architectural innovation is to use an external knowledgebase, based on RAG \citep{NEURIPS2020_6b493230}, and combine this seamlessly with a memory mechanism to improve the model's predictive performance. The main structure of this model is to use a question and document encoder, both transformers, to learn and look up passages of text from a knowledgebase (based on DPR; \citealt{karpukhin-etal-2020-dense}) and then fuse this knowledge into a transformer encoder/decoder model such as BART \citep{lewis-etal-2020-bart} or T5 \citep{Raffel2020ExploringTL}. Similar models including REALM \citep{Guu2020RetrievalAL},
Hard EM \citep{min-etal-2019-discrete}, SpanSeqGen \citep{min-etal-2020-ambigqa}, and Fusion-in-Decoder \citep{Izacard2021LeveragingPR} have achieved state-of-the-art results in factual domains such as answering natural language questions, trivia or games such as Jeopardy. In these domains, the key insight is that offloading knowledge externally allows models to perform better than much larger transformers that need to encode all knowledge in their weights. These methods that rely on retrieving raw text are also competitive with those that have tried to incorporate structured information such as GraphRetriever \citep{Min2019KnowledgeGT} or PathRetriever \citep{Asai2020Learning}. We experiment both with a \textit{Wikipedia} KB and \textit{Wikiplots}, a KB of story plot summaries. The motive for the latter is that these plot fragments or vignettes act as a planning system (or schema; \citealt{Schank1977ScriptsPG}) guiding expectations. \citet{Riedl2008StoryPW} used a similar concept in a rule-based system. \citet{sap-etal-2020-recollection} also use a bag-like episodic memory mechanism for inference in stories without the more sophisticated transformer encoders of the RAG model. After the experimental work in this paper, a follow-up paper by \citet{DBLP:journals/corr/abs-2104-07567} on several RAG variants found that the KB was able to reduce the amount of hallucination in generating dialogue. The KB grounds the text generation in relevant facts retrieved from the KB. While the story domain is different intuitively, the same effect is desirable; inferring salience should be grounded either in plot knowledge from \textit{Wikiplots} or general knowledge from \textit{Wikipedia}, and also the memory of the previous character actions and plot developments.

\section{Methods}

\subsection{Model}

\begin{figure*}[htbp]
\centering
\includegraphics[width=0.95\textwidth]{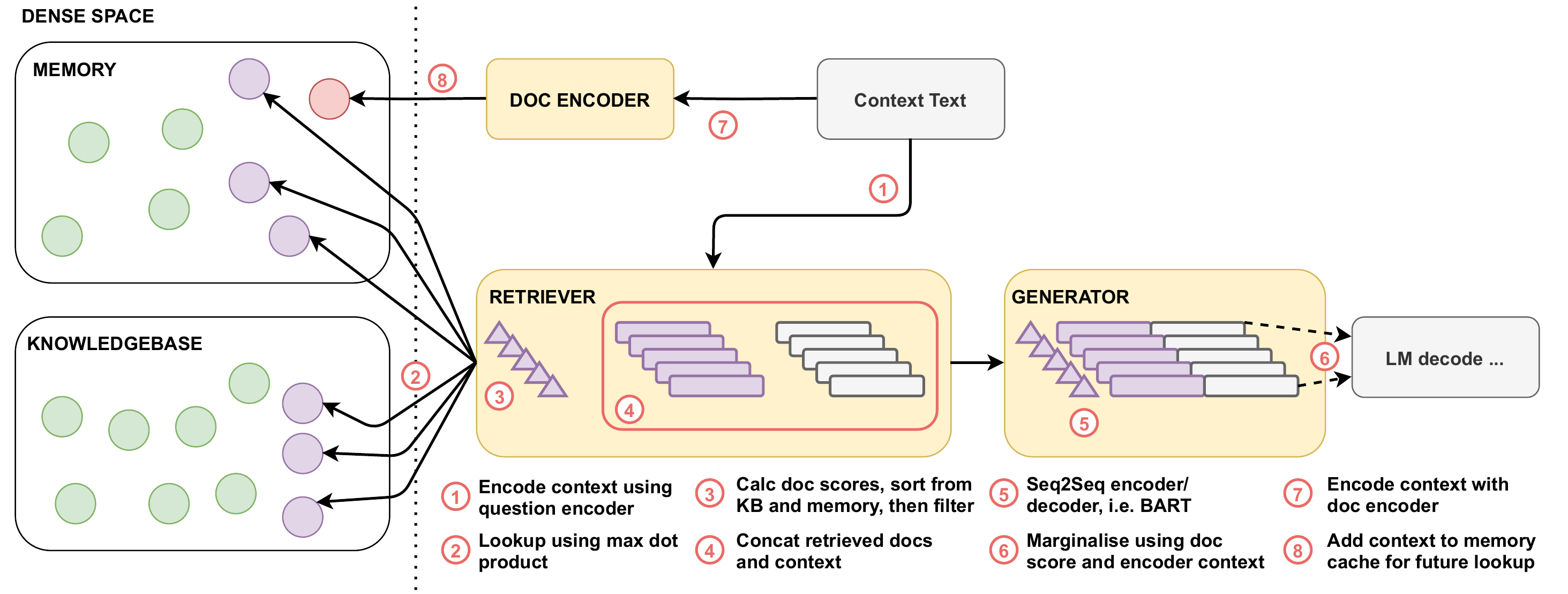}
\caption{Architecture of the memory RAG model: On the left-hand side are caches containing the permanent KB and transitory memory, which seen passages are added to. The Retriever encodes context text, looks up from both KB and memory, and concatenates the retrieved text to the context text. The generator, the BART encoder-decoder processes each passages concatenation, and marginalises over them to produce a distribution over output wordpieces. }
\label{fig:memory_rag}
\end{figure*}

The RAG model has been extended to incorporate a memory module, see Figure~\ref{fig:memory_rag}. \footnote{The code is provided via Github at \url{https://github.com/dwlmt/story-fragments/tree/emnlp2021-submission/}}. Seen passages are added to the memory cache (configurable FIFO or LRU). The model retrieves $n$ passages, performs a lookup in both the KB and memory and then reranks them together using the dot product score between the question and document encoder vectors.  A significant benefit is that it naturally integrates both a short-term and long-term KB retrieval mechanism with a relatively simple design while allowing a powerful pre-trained LM (BART Large; \citealt{lewis-etal-2020-bart}) and retrieval systems (DPR; \citealt{karpukhin-etal-2020-dense}) from RAG to be retained. For comparison, we train a \textit{baseline} model in which only the question encoder from RAG is finetuned so that existing KBs can be used without becoming stale. We also compare to the \textit{mem} model, where both the question and document encoder are finetuned, and only memory is used during inference.

The notation follows from the RAG paper and the model derived from the RAG-Token model. Assuming $x$ is the original input or context text, and $y$ is the target label for upcoming text, and $z$ a passage of text from a retrieved document from the KB or memory, $t$ a time index from the place of the passage in the story, and $\theta$ the parameterisation of the model. The generation task, $p_{ \theta }(y_{t} \mid x,z,y_{1:t-1})$, is to predict $y_{t}$ by marginalising over the input, previously generated word pieces and the retrieved document, this is defined in~(\ref{eqn:rag_token}). Each next token is marginalised over all the retrieved $z$ documents. The respective probability varies for each $z$ at each step for each retrieved passage.
\begin{equation}
\begin{split}
P(y \mid x)  &\approx \\  &\prod ^{N}_{t}  \sum_{z \in \text{Z}(p( \cdot \mid x))} p_{ \mu } (z \mid x)p_{ \theta }(y_{t} \mid x,z,y_{1:t-1})
\end{split}
\label{eqn:rag_token}
\end{equation}
The top $z \in Z$, by default five, passages are retrieved by maximising the dot product, $p_{\mu}(z \mid x) \propto \exp(\mathbf{d}(z)^{T}\mathbf{q}(x))$, where $\mathbf{d}$ is the document encoder, and $\mathbf{q}$ the question encoder, both pre-trained from the DPR \textit{multiset} model, resulting in a bi-encoder setup. Only $\mathbf{q}$ is finetuned in training.  The text passages $z$, whether retrieved from the KB or memory, are then concatenated onto the original $x$ text and fed through the BART large encoder/decoder model. The memory mechanism for training is a single pool of up to $128$k vectors that operates as an LRU cache during training.

The principal training loss in~(\ref{eqn:nll}) is simply the negative log likelihood over the batch as per the standard left-to-right decoder loss for BART. Because the model marginalises the retrieved passages, back-propagation through this loss also updates the question encoder to retrieve more relevant passages. 

\begin{equation}
 \mathcal{L}_{nll}(y) = \sum_{j} -\log \: p(y_{j} \mid p_{j})
\label{eqn:nll}
\end{equation}

\subsection{Training}

Datasets are read in an interleaved or round-robin fashion so that only one $(x,y)$ pair from each story is in a batch. Batches are sliding windows of $12$ sentences for both $x$ and $y$ with a $k$ of five passages to retrieve. The combined context for the concatenated encoder text is truncated to $512$ word pieces, and the max length for the decoder is $128$. The model is trained with a batch size of $32$. RAG has a delimiter separating retrieved text when concatenating for BART. We swap the order of RAGs concatenation so that the context is first and answer passages second to prevent truncation of the context text.

To allow the model to train on 12GB GPUs, we use the zero optimisation memory saving features of DeepSpeed \citep{10.1145/3394486.3406703}, which also necessitates using FP16, with gradient checkpointing for the model. Our training uses the base version of the RAG multiset encoders and the original pre-trained BART Large. We finetune with Adam \citep{DBLP:journals/corr/KingmaB14} with a learning rate of $2^{-6}$.

\subsection{Datasets}

\textit{BooksCorpus} \citep{Zhu2015AligningBA} provides a large corpus of longer novel-length works and is used for training. However, \textit{BooksCorpus} consists of free books scraped from \href{https://www.smashwords.com/}{Smashwords}; these works are highly slanted towards particular genres such as romance and fantasy which are unlike the evaluated task, which is mainly classic works. To supplement BooksCorpus an additional training dataset from Gutenberg using the \href{https://github.com/c-w/gutenberg}{c-w/gutenberg} library filtered to only English language fictional works. Another important area of longer-form storytelling is movies or dramatic works. So to improve diversity, the Movie Scripts datasets \citep{Ramakrishna2017LinguisticAO} is used. Multi-dataset models performed better on the validation set in training than single corpus models, so only these are evaluated. The training set sizes are \textit{BooksCorpus} circa $18$k works, \textit{Gutenberg} $27$k, and \textit{Movie Scripts} $1.5$k. We split sentences using \href{https://github.com/microsoft/BlingFire}{Blingfire}. 

\subsection{Baselines}

The primary baselines for salience prediction come from \citet{otake-etal-2020-modeling}. \textit{Random} just randomly assigns a salience score to each sentence position. \textit{Ascending} assigns scores that increase per position. \textit{Descending}, the reverse, assigns decreasing scores per position. The intuition behind these benchmarks is that important information can be clustered at the beginning or end of a story or chapter.

Otake et al. use TF-IDF as another benchmark; we use a BERT derived clustering summarisation approach \citep{Miller2019LeveragingBF}. The method uses k-means to cluster BERT sentence vectors according to the number of desired sentences and then selecting the sentences closest to the centroids. Since \textit{salience} scores are required, we adapt this method to output the cosine distance from the centroid as a salience score. We set the $k$ so that there is one cluster for every $10$ sentences. One change from Miller is to use the \textit{stsb-roberta-large} sentence transformers model \citep{reimers-2019-sentence-bert}, which has sentence embeddings that perform much better on a range of semantic tasks than raw BERT. 

\subsection{Inference}

\begin{figure*}[htbp]
  \centering
  \includegraphics[width=0.77\textwidth]{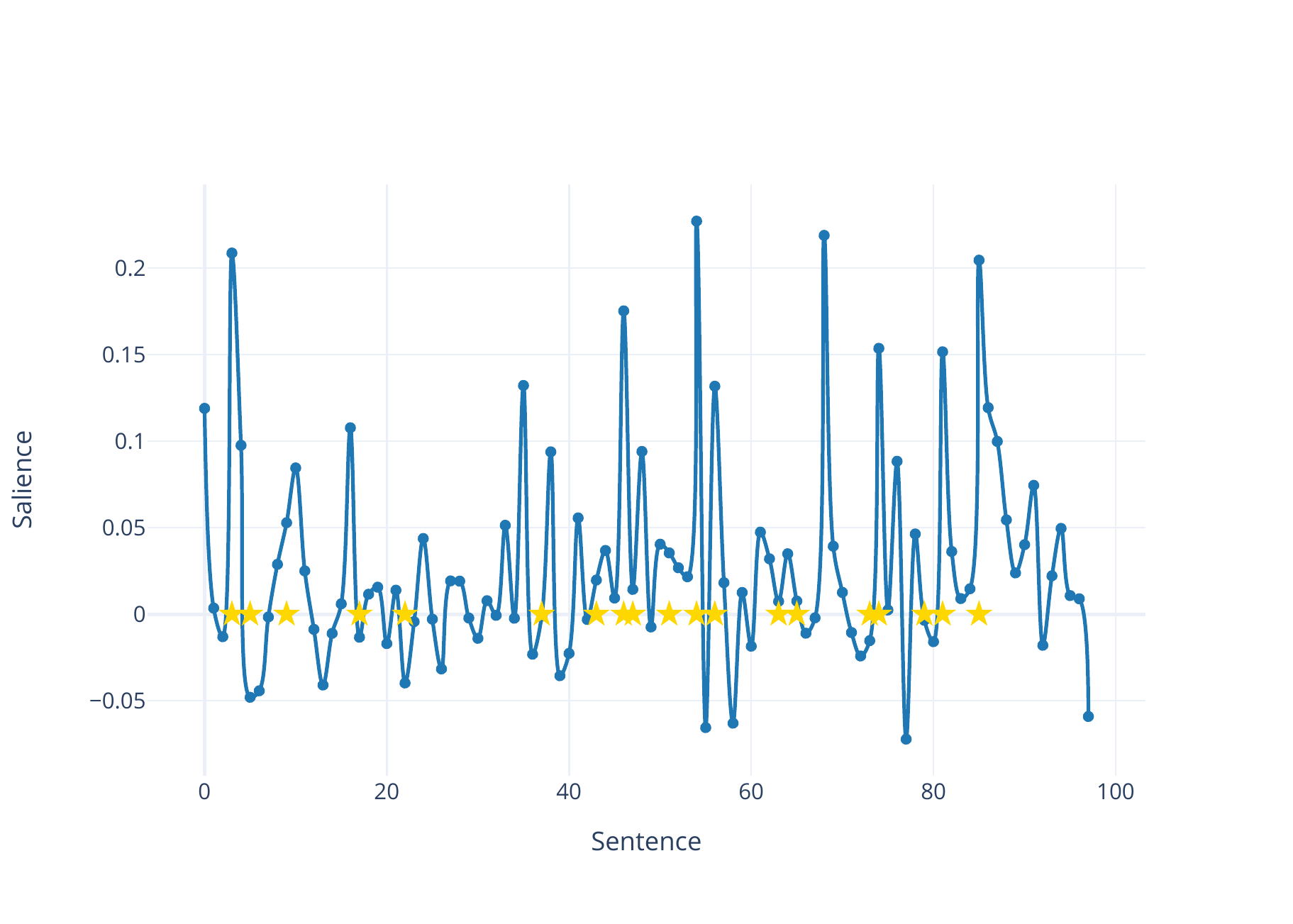}
  \caption{The Like-Sal of \textit{Moby Dick Chapter 1}. Stars are the Shmoop labels. Further interactive examples in \textit{supplementary material}.}
  \label{fig:salience_plot}
\end{figure*}

\textit{Salience} detection is based on the BCF method \citep{otake-etal-2020-modeling}. We only use the sentence deletion variant. Let $S$ be the set of all sentences. The \textit{salience} is $\sigma$. For BCF this uses an event removal function $r$ and coherence evaluator $c$. $c$ is the difference in coherence between when the sentence $t$ is present and removed in (\ref{eqn:del}) for the following $n$ sentences. Note that $r$ can be used more broadly as a structural manipulation function. In this paper $r$ is also used for \textit{swap} function and a \textit{knowledge difference} function, these are described later.
\begin{equation}
    \sigma(S_t,S_{\{1:n\}}) = c(S_{\{1:n\}}) - c(\tilde{S}_{\{1:n\}})
\label{eqn:del}
\end{equation}
The coherence (\ref{eqn:coh}) and (\ref{eqn:coh2}) is the average log-likelihood of the word pieces following sentences up to the maximum word pieces of the label, normalised by the length~(\ref{eqn:z}).
\begin{equation} 
\begin{split}
    &c(S_{\{1:n\}}) = Z  \log P^{}_{}(S_{\{t+1:n\}} \mid S_{\{1:t-1\}}, S_{t}) 
\end{split}
\label{eqn:coh}
\end{equation}
\begin{equation} 
\begin{split}
    &c(\tilde{S}_{\{1:n\}}) = Z \log P^{}_{}(S_{\{t+1:n\}} \mid S_{\{1:t-1\}}, r(S_t))
\end{split}
\label{eqn:coh2}
\end{equation}
\begin{equation} 
\begin{split}
    Z = \frac{1}{\lvert S_{\{t+1:n\}} \rvert} 
\end{split}
\label{eqn:z}
\end{equation}
We treat a sentence as an event. In inference, we use a context of $12$ sentences (truncated to $512$ word pieces) and up to $20$ passages are retrieved either from the KB or memory. Otake et al. run \textit{salience} from each deleted sentence to the end of the story, which is factorial complexity for the number of sentences. This is infeasible on novel-length works, so our \textit{salience}  implementation is more localised and run over the next $128$ LM wordpieces.

As well as BCF Salience, several other measures for \textit{salience} are explored. We experiment with \textit{knowledge salience}, which measures the difference between \textit{salience} with the RAG KB and Memory enabled versus with it disabled. \textit{Swap salience} follows the same structure as sentence deletion, but the $r$ function swaps the order of the sentences rather than deleting them, and so tests order dependence as a form of salience. The sentiment is another relevant factor in whether something is salient; more emotional passages, either negative or positive, might be more salient. We use VADER sentiment \citep{Hutto2014VADERAP} as an adjustment factor for other \textit{salience} measures $salience \cdot (1.0 + \abs(sentiment))$ where sentiment is the absolute values of the sentiment in the range $0.0-1.0$. In addition, we follow \citet{wilmot-keller-2020-modelling} and define measures based on embeddings: We define $E$ as the average of the word piece vectors from the BART encoder after marginalisation. The first measure is the cosine distance from subsequent vectors, defined by Wilmot and Keller as \textit{Ely surprise} $cos\_dist(E_{t},E_{t-1})$. The second measure takes a vector distance rather than average log-likelihood in the sentence deletion BCF method to create a version based on an embedding, not LM decoding. The evaluated measures are:
\begin{itemize}
  \item \textbf{Clus-Sal}: The clustering baseline.
  \item \textbf{Like-Sal}: The main BCF measure described.
  \item \textbf{No-Know-Sal}: The same but with both the memory and KB disabled as per Otake et al.
  \item \textbf{Like-Imp-Sal}: Use sentiment to adjust the salience. 
  \item \textbf{Like-Clus-Sal}: Combining the \textit{Like-Sal} and {Clus-Sal} measures via weighted addition: $\text{Clus-Sal} + 2 * \text{Like-Sal}$.
  \item \textbf{Like-Clus-Imp-Sal}: Additionally adjusting for the impact sentiment.
  \item \textbf{Know-Sal}: The difference between average log-likelihood of the LM with the KB and memory on versus off, \textit{knowledge salience}.
  \item \textbf{Swap-Sal}: Use the same BCF approach but swaps rather than deletes a sentence to test structural ordering.
  \item \textbf{Emb-Surp}: The \textit{Ely surprise} cosine distance measure.
  \item \textbf{Emb-Sal}: \textit{Salience} based on above embedding distance not average log-likelihood.
\end{itemize}
We run the evaluation on three models: With the \textit{Wikiplots} dataset, with the \textit{Wikipedia} dataset, and with just \textit{mem} enabled and additional finetuning of the document encoder.

\section{Experiments}

\subsection{Perplexity Model Improvements}

The major innovation of the RAG derived model is incorporating the KB and memory mechanism into the LM, and therefore, it needs to be tested what impact it has as a general LM. Table~\ref{tab:perplexity} shows the baseline model median perplexity with combinations of KB and memory access turned off.

\begin{table}[htbp]
\centering
\begin{tabular}{@{}lc@{}}
\toprule
\textbf{Model}     & \textbf{Perplexity $\downarrow$} \\ \midrule
LM+Mem+KB & 19.44         \\
LM+KB & 19.37         \\
LM+KB(Wikipedia) & 19.94 \\
LM+Mem & \textbf{15.95}         \\
LM & 66.00         \\
LM+Scram(Mem+KB) & 60.21 \\

\bottomrule
\end{tabular}
\caption{Median perplexity of the baseline model. Plus means that type of memory or KB is enabled. Scram means that random passages have been retrieved from the KB and memory. Wikiplots KB unless Wikipedia is specified. All over the first five stories in the dataset using $20$ retrieved passages.}
\label{tab:perplexity}
\end{table}

The best model is the baseline with only the memory, and the KB turned off. Both versions of the KB on their own and memory combined are slightly worse and around the same perplexity. The crucial difference is that \textit{LM}, the model with neither, is far worse, and scrambling, which retrieve random passages, is only slightly better. Overall, these results validate that memory and KB are hugely improving the predictive power of the LM.

\subsection{ProppLearner}

Following on from the BCF paper \citep{otake-etal-2020-modeling}, we evaluate the ProppLearner task derived from the Propp dataset \citep{Finlayson2017ProppLearnerDA}, a richly annotated corpus of 15 Russian fairytales translated into English. See Otake et al. for more rationale for the link, but the Proppian functions with which this corpus is annotated define stereotypically important roles in the classic Russian fairytale. They represent the key events of a story's plot, which should therefore be salient. As per Otake et al., the results are reported using MAP (mean average precision; \citealt{Manning2008IntroductionTI}).

\begin{table}[htbp]
\centering
\begin{tabular}{@{}lc@{}}
\toprule
\textbf{Model}     & \textbf{MAP $\uparrow$} \\ \midrule
Random             & .213         \\
Ascending          & .277         \\
Descending         & .185         \\
TF-IDF             & .279         \\ \midrule
Otake Sal       & .280         \\
Otake Comb Sal   & .301         \\ \midrule
Clus-Sal & .275         \\
Like-Sal & \textbf{.319}         \\
Like-Imp-Sal & .313         \\
Know-Diff-Sal & .309         \\
Swap-Sal & .236  \\
Emb-Surp & .247         \\
Emb-Sal & \textbf{.319}  \\\bottomrule
\end{tabular}
\caption{Compare Otake et al. baselines and models with RAG equivalents. }
\label{tab:propp_table}
\end{table}

 All the RAG models are the baseline used with different variants of the \textit{salience} measures. The best RAG models, see Table \ref{tab:propp_table}, measures \textit{Like-Sal} and \textit{Emb-Sal} score slightly better than Otake et al.'s model. This validation is limited, though, as the Propp dataset is tiny, with only $15$ stories of less than $150$ sentences and limited annotations. In the next section, we extend this evaluation approach to a corpus of much longer works of classical literature, using silver labels derived from a corpus of aligned summaries. This allows us to test both the memory and KB mechanism adequately, as these would be expected to be most advantageous for longer works. This approach will enable us to test whether our method scales beyond short texts and adds robustness to the evaluation through the breadth of the corpus and the challenging nature of the text.

\subsection{Shmoop Automated Evaluation}

\begin{table*}[htbp]
\centering
\begin{tabular}{@{}lccccccccc@{}}
\toprule
 \multicolumn{1}{c}{\textbf{Measure}}                    & \multicolumn{3}{c}{\textbf{MAP $\uparrow$}}    & \multicolumn{3}{c}{\textbf{Rouge-L $\uparrow$}} & \multicolumn{3}{c}{\textbf{Recall K $\uparrow$}} \\ \midrule
                                     \multicolumn{1}{l}{}                  & \textbf{Plots} & \textbf{Mem} & \textbf{Pedia} & \textbf{Plots}  & \textbf{Mem} & \textbf{Pedia} & \textbf{Plots}  & \textbf{Mem}  & \textbf{Pedia} \\ \midrule
                Random                                & .178           & .178         & .178           & .250            & .250         & .250           & .132            & .132          & .132           \\
 Ascending                             & .152           & .152         & .152           & .243            & .243         & .243           & .163            & .163          & .163           \\
Descending                            & .207           & .207         & .207           & .180            & .180         & .180           & .109            & .109          & .109           \\
 Clus-Sal                              & .230           & .230         & .230           & .296            & .296         & .296           & .187            & .187          & .187           \\ \midrule

 No-Know-Sal                           & .246           & .246         & .246           & .336            & .336         & .336           & .205            & .205          & .205           \\ \midrule

 Like-Sal                              & .294           & .280         & .288           & .368            & .356         & .359           & .254            & .241          & .243           \\
                                   
                                Like-Imp-Sal                          & .291           & .276         & .287           & .367            & .352         & .369           & .253            & .238         & .251           \\
                                Like-Clus-Sal                         & .291           & .273         & .287           & .355            & .339         & .358           & .245            & .228          & .243           \\
                              Like-Clus-Imp-Sal & .289           & .276         & .285           & .351            & .336         & .355           & .240            & .225          & .251           \\
                                  Know-Diff-Sal                         & .246           & .242         & .243           & .301            & .300         & .306           & .199            & .194          & .200           \\
                                  Swap-Sal                              & .256           & .241         & .252           & .309            & .294         & .313           & .210            & .193          & .210 \\
 Emb-Surp                              & .249           & .196         & .243           & .311            & .315         & .315           & .201            & .245          & .200  \\
 
  Emb-Sal                              & \textbf{.312}           & \textbf{.311}         & \textbf{.309}           & \textbf{.413}            & \textbf{.371}         & \textbf{.419}           & \textbf{.271}            & \textbf{.271}          & .\textbf{272} \\

  \bottomrule
\end{tabular}
\caption{Shmoop results from the silver label evaluations.}
\label{tab:shmoop_results}
\end{table*}

Ideally, to evaluate this thesis on longer works, there would be a set of Gold standard annotations with the salient sentences. Typically even short novellas can be over $20$K words, more normal novels longer than $50$K words. More sweeping works such as \textit{Anna Karenina}, \textit{Wuthering Heights}, \textit{The Fellowship of the Ring}, or \textit{David Copperfield} can be well over $100$K words. Per-sentence annotations for longer works such as novels and plays are prohibitively expensive. This is especially true when multiple annotators are required to ensure high inter-annotator agreement. It would also not be possible with insufficiently trained and lower cost crowdsourced workers. Reading a local passage would not be enough as it is only possible to judge \textit{salience} over the whole narrative, which can be tens of thousands of words. This requires strong comprehension and thus requires skilled annotators and is a daunting annotation task. Instead, this paper builds on a variant of an approach for event \textit{salience} in news articles \citep{liu-etal-2018-automatic-event,jindal-etal-2020-killed}. The method is to align expert-written summaries with the full text, tagging sentences that align with the summary as salient, thus turning the evaluation into a binary ranking problem. The intuition is that the summary will mention only salient events and themes.

We use the \href{https://github.com/achaudhury/shmoop-corpus}{Shmoop corpus} \citep{DBLP:journals/corr/abs-1912-13082}, which contains classic works of literature, such as \textit{Moby Dick}, but also plays such as \textit{A Midsummer Nights Dream}, and short stories including \textit{The Mask of the Red Death}.  The Shmoop corpus has stories split into chapters with aligned summaries. These bullet point summaries, if colloquial in style, are professionally written as study guides for students. They are written with a deep understanding of the plots and the salient events in them, which can serve as a valid proxy for salience. Conceptually they are also similar to the \textit{ProppLearner} evaluation, although without specific Proppian roles, which are unused anyway for binary \textit{salience} classification. It also aligns with the BCF concept, as if events from the summary are removed, they would significantly alter the plot.\footnote{There are occasional exceptions, such as summary points that discuss themes of the overall work and not specific plot events, but these are rare.} \citet{jindal-etal-2020-killed} align summaries to text by using BERT \citep{Devlin2019BERTPO} to match constituent parts of events extracted from semantic role labels (SRLs). However, in testing, this performed poorly. Unlike news, the story summaries are more loose descriptions of events, which the SRL method struggles with. We instead found using an S-Bert transformer \citep{reimers-2019-sentence-bert} on the whole sentence worked much better in aligning summaries to the full text. The method is as follows:\footnote{Full examples are referenced in the \textit{supplementary material}.}
\begin{enumerate}
  \item Split aligned chapters into sentences, $S_t$ for summaries and $F_t$ for the full text.
  \item Extract sentence embeddings using the \href{https://www.sbert.net/}{Sentence Transformers} model \textit{stsb-roberta-large} \citep{reimers-2019-sentence-bert} , $r(S_t)$ and $r(F_t)$ .
  \item Calculate cosine similarity for all pairwise $r(S_{t})$ and $r(F_{t})$ for $t \pm \rho$, where the range is $\rho = 10.0\%$ and the valid range for $t$ is $x \in \mathcal{X}$ for $S_{t}$, and $y \in \mathcal{Y}$ for $F_{t}$.
  \item Mark up to $k$ as salient sentences for all sentence pairs in the alignment window $s(x, y) =cos\_sim(r(S_{t_x}),r(F_{t_y})$ where:
  \begin{itemize}
    \item $k = 3$
    \item $s(x, y) \ge \mu$, $\mu = 0.35$
    \item $s(x, y) \ge  \argmax_{x \in \mathcal{X}, y \in \mathcal{Y}} s(x, y) - \theta$, where $\theta = 0.05$
  \end{itemize} 
\end{enumerate}
Pairs of summary and full-text sentences are matched within a percentile range. The rationale is that matches are likely to occur in the full text in a roughly similar position to the summary.  We allow up to three targets per summary sentence, as the summary sentences often compress information with multiple clauses and because sometimes there are near identically suitable matches. The advantage of this method is that it allows automated evaluation of \textit{salience}  to scale to longer works that test the memory and KB mechanism of the model without excessive annotation cost. The silver Shmoop annotations are on $226$ titles, spanning $6,939$ chapters with $214,617$ silver standard labels. Each chapter averages $148$ sentences with an average of $31$ labelled as salient using the criteria specified. See Figure~\ref{fig:salience_plot} for an example of the Shmoop labels plotted with the salience for a book chapter.

As for the ProppLearner data, we report MAP. We also evaluate with ROUGE-L \citep{lin-2004-rouge}, comparing the text by selecting the $k$ most salient sentences according to the measure where $k$ is the number of salient sentences, and report recall at $k$. All measures are calculated by chapter, and we take the mean across the dataset.

The results in Table~\ref{tab:shmoop_results} reveal several main themes. The \textit{Clus-Sal} baseline measure improves on all the other baselines but only by a comparatively small margin with the best of each, by $0.03$ compared with the best MAP baseline, $0.04$ with Rouge-L, and $0.02$ with recall. The baseline is a centroid based extractive summarisation model that uses a powerful transformer; the relatively small performance improvement increase shows that the task is challenging.

The main \textit{Like-Sal} measure shows an improvement of around $0.05$ over \textit{Clus-Sal}, and  $0.10$--$0.15$ over the baseline. This is a reasonable improvement given the model is unsupervised. The \textit{No-Know-Sal} (without memory and KB) is about $0.03$--$0.04$ worse on MAP and recall, which indicates that the RAG enhancements are helping improve \textit{salience} detection. The theoretical reason would be that BCF detects shifts in state and the informed model with the KB and memory is more likely to predict more obvious events. So salient events are more likely to be significant plot shifts. The biggest finding is that \textit{salience} based on the embedding, \textit{Emb-Sal} is the strongest measure. This shows the merit of using the BART model more flexibly as a general-purpose sentence encoding model. The \textit{Emb-Surp} measure is a slight improvement on the baselines, indicating that it is mainly the BCF method that causes an improvement in \textit{salience} detection, rather than a simple measure of how much the story changes from sentence to sentence.

One difference from the Otake et al. finding is that combining the \textit{Clus} measures makes little difference. Neither do the \textit{Imp} measures that use absolute sentiment score. While worth exploring further this is consistent with \citet{wilmot-keller-2020-modelling} findings when adjusting sentiment with inferring \textit{surprise} and \textit{suspense}.

Of the more esoteric measures, both \textit{Swap-Sal} and \textit{Know-Sal} improve on the baseline, although not by much. The more interesting is \textit{Know-Diff-Sal}, which performs similarly to the \textit{Clus-Sal} baseline. The measure as a proxy to exploit the difference between reader and character is quite crude. There may be a more sophisticated way to develop this idea by modelling character knowledge explicitly.

Largely speaking, there does not seem to be much of a difference between the different memory and KB configurations. With the best measure \textit{Emb-Sal}, the results are nearly identical. With the original BCF measure \textit{Like-Sal} and its variants, both the \textit{Wikiplots} dataset (plot summaries from Wikipedia) and the full \textit{Wikipedia} dataset only result in a tiny improvement. It might be expected that a KB would improve performance for salience prediction, but recall that in the perplexity evaluation, memory-only performed better. The present results also suggest that the memory mechanism is the main reason for the improvement over \textit{No-Know-Sal}.

The memory and KB  access pattern of the model is highly non-linear and references the earlier mention of the same characters, places, or moods. One example of this is from \textit{Great Expectations} final chapter, where \textit{Pip} and \textit{Estella} have their last meeting. The passage most recalled is their early meeting some $100$K odd words earlier while walking in the Garden where \textit{Estella} plays a trick on \textit{Pip}. The memory focuses on the characters and their relationship rather than many irrelevant details and subplots occurring in between. The episodic memory can be thought of as acting as an index into crucial elements of the plot, which is essential for narrative comprehension \citep{Zwaan1999FiveDO,Zwaan1995TheCO}. It justifies the suitability of an episodic memory model for understanding longer-form narrative texts.

\section{Conclusion}

The main overall finding is that the BCF method can infer \textit{salience} over and above baselines with an improvement on much longer works. We find that augmenting an LM with memory and an external KB can improve the detection of salience and increase the predictive power of the LM on narrative text. We also find that a vector-based version of the concept can perform slightly better than using the log-likelihood from an LM. Therefore, this paper demonstrates that it is feasible to run an unsupervised method on novels from Dickens or plays by Shakespeare and achieve correlation with an automated silver label benchmark. Nevertheless, the MAP results are around $0.3$, and ROUGE-L is $0.4$, which leaves room for improvement. 

One factor in the moderate increase could be that the \textit{salience} modelling is explicitly local over the label of the $n$ next tokens. This is more a local view of \textit{salience} as intended from the reader perspective. The model may flag up false leads that are locally important but not globally for the plot. In contrast, the \textit{Shmoop} is written with the knowledge of the whole story, and so will exclude them. A more reader orientated evaluation is for future work. Although the \textit{Shmoop} alignment is generally strong, there are occasions where arguably multiple sentences could be deemed the correct one, and the silver label is one and the \textit{salient} peak the other. With this unsupervised approach, performance is likely to be underestimated as the labels are entirely independent. In contrast to much recent supervised work, such as  PEGASUS \citep{Zhang2020PEGASUSPW} summarization \textit{Has additional human evaluation on some datasets.} system, use silver labels created with proxies such as ROUGE.  The labels both train the system and are evaluated on. Even with a separate test set performance, the system is more likely to replicate any noisy misalignments in the labelling process and overestimate performance.


On future work, if RAG can improve LM prediction performance and help infer salience, then the same models would seem to hold promise in improving text generation, including story generation. 

The \textit{knowledge salience} approach is a simple attempt to model the informed reader versus the naive one.  In narratology, the characters perspective is crucial in for example  \textit{eventfulness} \citep{schmid2003narrativity,Schmid2017EventfulnessAR}; \citet{Lotman1977TheSO} notion of characters \textit{crossing a forbidden semantic border}; or \textit{suspense} as per \citet{zillmann1996psychology}, or
\citet{Gerrig1994ReadersAP} concept of the \textit{reader as problem solver}. There is, therefore, rich potential work in modelling character states, knowledge, intents and contrasting them with the readers' expectations, and the norms of the narrative world in inferring concepts such as \textit{salience}, \textit{suspense}, and \textit{surprise}. Characters could be implicitly modelled using a per entity memory model extending the current RAG approach. Or take a more structured approach inspired by recent work such as \citet{sims-bamman-2020-measuring} modelling literary character communication, or story generation systems such as CAST  \citep{DBLP:journals/corr/abs-2105-01311} that model multiple characters goals or C2PO \citep{Ammanabrolu_Cheung_Broniec_Riedl_2021} that more explicitly models causal chain relations.

\section*{Acknowledgments}

The authors would like to thank the anonymous reviewers. We would also like to thank \textit{Shmoop} for permission to use and publish from the summaries. Wilmot's work is funded by an EPSRC doctoral training
award.

\bibliography{references}
\bibliographystyle{acl_natbib}

\clearpage

\appendix

\section{Examples}
\label{sec:appendix_examples}

\subsection{Interactive Plots}

Within \textit{supplementary material} are \textit{shmoop\_iteractive\_plots}. This is a selection of interactive plots \textit{all.html} from the Shmoop alignment of chapters. The plots show all the reported metrics that can be toggled using the legend. Each plot has the full text of the chapter with a sentence on each data point. The gold stars are the original Shmoop summary sentence, which sentence it aligns with, and the similarity score. Also included with each chapter is a correlation heatmap plot showing the Spearman $\rho$ correlation between all the reported metrics.

\subsection{Retrieved Passages}

To illustrate how the retrieval mechanism functions in \textit{supplementary material} within \textit{shmoop\_retrieved\_passages} are chapters from the end or near the end of \textit{Kim}, \textit{Great Expectations} and \textit{20,000 Leagues Under the Sea}. These json files contain a list of the distinct passages for the chapter of each book. For each passage there is a list of retrieved passages that were looked up from the KB and memory for each block, $10$ per passage, using the baseline. Each passage has a dot product score and the marginalised probability. The memory\_id for a passage indicates the relative position  in the story. The main reason for inclusion is it shows the memory lookup is highly non-linear and the retrieved passages from earlier in the story are strongly related to the characters, places and themes involved from anywhere in the story and not the most recent chapters. 

\subsection{Shmoop Alignment}

For Shmoop alignment some examples are also found in table \ref{tab:saliency_alignment_short} that illustrates a few different types of sentences and the matches against the full text. In the \textit{supplementary material}file within the \textit{shmoop\_alignments} folder are the alignment files for \textit{20,000 Leagues Under the Sea} and \textit{Richard III}. Only two are included to demonsrate the alignment since \href{https://github.com/achaudhury/shmoop-corpus}{Shmoop} requires permission to use the summaries for the dataset. The format of the file is:

\begin{itemize}
   \item Within the main alignment json files the main json element with the annoations is \textit{chapters} which splits books or plays into sub-sections. 
   \item Within each element of \textit{chapters} there is a \textit{summary} element. 
   \begin{itemize}
     \item \textit{summary} is a list with each of the summary sentences each one has a list of \textit{alignments} that contains the index, text, and cosine similarity score of the full text sentences it is linked to \footnote{The threshold in the file 0.3 for alignment. Within the evaluation script a tight filter is used on these alignments of 0.35.}. 
   \end{itemize}
    \item Within each element of \textit{chapters} there is a \textit{full\_text} element. 
   \begin{itemize}
     \item \textit{full\_text} contains a list of the sentences with the full text. Each sentence has a salient boolean attribute and a salience\_score attribute.
     \item Not included in the submission but these are exported to separate per book json files for running through the RAG model.
   \end{itemize}
\end{itemize}

For running the \textit{Shmoop} alignment from the Github repository:

 \begin{itemize}
     \item \textit{align\_shmoop\_summaries.py} processes the separate raw summaries and full text books into a single jsonl file.
     \item \textit{salience\_event\_processing.sh}, a slurm script runs the Shmoop alignment and produce runnable files for the model input. The file has configurable parameters for changing the thresholds and the models used.
   \end{itemize}

\begin{table*}[t]
\centering
\begin{tabular}{@{}lll@{}}
\toprule
\textbf{Summary}                                                                                                                                                                                                                                             & \textbf{Full Text}                                                                                                                                                                                                                                                                                                                                                                                                                   & \textbf{Score} \\ \midrule
\begin{tabular}[c]{@{}l@{}}Then, two huge columns \\ of water shoot from it, \\ knocking the crew down.\end{tabular}                                                                                                                                         & \begin{tabular}[c]{@{}l@{}}The electric light suddenly went out, and two\\ enormous waterspouts crashed onto the deck \\ of the frigate, racing like a torrent from stem \\ to stern, toppling crewmen, breaking spare \\ masts and yardarms from their lashings.\end{tabular}                                                                                                                                                       & 0.723          \\ \midrule
\begin{tabular}[c]{@{}l@{}}He grabs the extinguisher \\ cap thing and tries to smother\\ the kid/grandpa ghost with it.\end{tabular}                                                                                                                         & \begin{tabular}[c]{@{}l@{}}In the struggle, if that can be called a struggle \\ in which the Ghost with no visible resistance \\ on its own part was undisturbed by any effort\\ of its adversary, Scrooge observed that its light \\ was burning high and bright; and dimly \\ connecting that with its influence over him, he\\ seized the extinguisher-cap, and by a sudden \\ action pressed it down upon its head.\end{tabular} & 0.600          \\ \midrule
\begin{tabular}[c]{@{}l@{}}Sara wants to say that she\\ already knows French, but \\ she doesn't know how to say\\ so and ends up giving Miss \\ Minchin the impression that\\ she's being difficult and doesn't \\ want to learn the language.\end{tabular} & \begin{tabular}[c]{@{}l@{}}Miss Minchin was a very severe and imposing\\ person, and she seemed so absolutely sure that\\ Sara knew nothing whatever of French that \\ she felt as if it would be almost \\ rude to correct her.\end{tabular}                                                                                                                                                                                        & 0.722          \\ \midrule
\begin{tabular}[c]{@{}l@{}}Emerson still feels rough\\ about ruining the lecturer’s talk \\ in the chapel.\end{tabular}                                                                                                                                      & \begin{tabular}[c]{@{}l@{}}But Mr. Emerson, contrite and unhappy, \\ hurried away to  apologize to the \\ Rev. Cuthbert Eager.\end{tabular}                                                                                                                                                                                                                                                                                          & 0.486          \\ \midrule
\begin{tabular}[c]{@{}l@{}}She thinks Dinah should find a \\ nice man and settle down.\end{tabular}                                                                                                                                      & \begin{tabular}[c]{@{}l@{}}And then you might get married to some decent\\ man, and there'd be plenty ready to have you,\\ if you'd only leave off that preaching, as is\\ ten times worse than anything your\\ Aunt Judith ever did.\end{tabular}                                                                                                                                                                                                                                                                                          & 0.334          \\ \midrule

\begin{tabular}[c]{@{}l@{}}According to him, the driftwood\\ is dry and ideal for starting a fire.\end{tabular}                                                                                                                                      & \begin{tabular}[c]{@{}l@{}}It is now dry and would\\ burn like tinder.\end{tabular}                                                                                                                                                                                                                                                                                          & 0.630          \\ \midrule

\begin{tabular}[c]{@{}l@{}}Detectives were sent to each port\\ in England to see if the money might \\be recovered.\end{tabular}                                                                                                                                      & \begin{tabular}[c]{@{}l@{}}As soon as the robbery was discovered, picked \\detectives hastened off to Liverpool, Glasgow,\\ Havre, Suez, Brindisi, New York, and other ports,\\ inspired by the proffered reward of two thousand \\ pounds, and five per cent. on the sum that might\\ be recovered.\end{tabular}                                                                                                                                                                                                                                                                                          & 0.618          \\ 

\bottomrule
\end{tabular}
\caption{Example showing the alignment of summary with full-text sentences, the score is cosine similarity. The examples are all chosen because the previously used SRL event extraction fails with this approach and matches incorrect sentence and the examples show different strengthes of matches and types of sentence.}
\label{tab:saliency_alignment_short}
\end{table*}

\section{Environment and Reproducibility}

\subsection{Setup}

The environment can be setup either via the \textit{requirements.txt} file with pip on the Anaconda \textit{environment.yaml} file, both in the Github repository. 

\subsection{Training Datasets}

The datasets are: \textit{BooksCorpus}, \textit{Gutenberg}, \textit{ScriptCorpus}, and \textit{Wikiplots} (as a KB, and not for training).

The preprocessing for all datasets is the same:

\begin{enumerate}
  \item Sentence splitting using \href{https://github.com/microsoft/BlingFire}{Blingfire}.
  \item Stories are randomly shuffled according to a fixed seed.
  \item There is a 80/10/10 training/validation/test split but this is only used for early stopping in training since evaluation is on separate datasets - ProppLearner and Shmoop. 
\end{enumerate}

\subsection{Training and Inference}

\begin{itemize}
  \item \textbf{Training}
    \begin{itemize}
        \item{Config}: The config files are in the \textit{training\_config}. The reported models are the 12 sentence block variants without experiment entmax or unlikelihood training which isn't reported.
        \item \textbf{Models}: The model files will be made available via Github when the anonymity period ends. BART Large has $400M$ parameters, the question encoder has $110M$ parameters and the doc encoder has $110M$ parameters. In the baseline model all apart from the doc encoder is finetuned, and all are finetuned with the memory only models.
        \item \textbf{Policy}: All models were trained with batch size $20000$ instance per epoch in training and $2000$ for validation. The early stopping policy is to stop training after $3$ epochs without improvement.
        \item \textbf{Time}: For the baseline model training took 9 days. Other models are comparable.
        \item \textbf{Epochs}: Baseline model training ran for 11 epochs, again other models are similar.
        \item \textbf{Validation Performance}: The best validation loss is $398$ (sum of the log likelihood) from $694$ on an untrained model.
    \end{itemize}
  \item \textbf{Inference}
     \begin{itemize}
     \item{Computation}: Inference computation depends on which salience measures are enabled. The main salience BCF method requires two passes through the text. Adding in knowledge or swap salience adds another pass for each. This is because the text must be passed through with an without each structural change. With all methods enabled for long works such as \textit{The Brothers Karamazov}, \textit{Emma}, \textit{Moby Dick}, or \textit{Great Expectations} all $>150K$ words inference time is typically $4-6$ hours running on a single GPU. This is pretty reasonable given the length of the works, and obviously much shorter novels and plays have proportionally shorter inference time. 
     \item{Memory}: The base configuration uses $28GB$ of general purpose memory, this needs to be increased to $64$ is the full Wikipedia KB with $23M$ passages is used.
     \end{itemize}
\end{itemize}

\subsection{Evaluation}

Within the Github repository the main evaluation scripts is \textit{salience\_evaluation.sh}. This script produces a per chapter csv file with all the evaluation metrics stats, and a single aggregated whole.
It used evaluating output in jsonl format produced by  \textit{predictor\_batch.sh} the main script to run salience inference with an existing model over a batch of stories. There is also a \textit{salience\_plot.sh} script for producing the interactive charts for each evaluation output.

Note more documentation is needed for the env variables to set but they are fairly self-explanatory in the slurm files. The main inference code is in the \textit{story\_fragments/predictors} package. The config is largely read through env variables in the python script. These need to be documented further for a full code release.

\end{document}